\documentclass[sigconf]{acmart}
\AtBeginDocument{%
  }

\usepackage{amsfonts}
\usepackage{amsmath}
\usepackage{array}
\usepackage{enumitem}
\usepackage{booktabs}
\usepackage[dvipsnames]{xcolor}

\usepackage{graphicx}
\usepackage{subcaption}
\usepackage{algorithm}
\usepackage{algorithmic}
\usepackage{float}

\setcopyright{none}
\renewcommand\footnotetextcopyrightpermission[1]{}
\acmConference{}{}{}
\acmBooktitle{}
\acmDOI{}
\acmISBN{}

\begin{document}

\title{Learning from Human Corrections through Temporal Preferences}

\author{Anjiabei Wang}
\orcid{0009-0008-4345-333X}
\affiliation{
  \institution{Yale University}
  \city{New Haven, CT}
  \country{United States}}
\email{anjiabei.wang@yale.edu}

\author{Tesca Fitzgerald}
\orcid{0000-0003-0867-0546}
\affiliation{
  \institution{Yale University}
  \city{New Haven, CT}
  \country{United States}}
\email{tesca.fitzgerald@yale.edu}


\begin{abstract}
    When people intervene to correct a robot, they provide information not only through \emph{how} they correct its behavior, but also through \emph{when} they choose to intervene. Existing approaches to Learning from Corrections focus on comparing the new trajectory to the robot's originally intended behavior, thus underutilizing the context of the robot's behavior \emph{before} the intervention. In this work, we investigate how both pre- and post-intervention behavior can be translated into preference pairs for learning value functions that approximate a human's latent reward. We construct post-intervention preferences by comparing corrected behavior with the behavior it replaces, and pre-intervention preferences from temporal relationships among states leading up to an intervention. We evaluate the learned value representations through two downstream tasks: goal inference and correction forecasting. Our results show that post-intervention preferences are particularly effective for goal inference, while pre-intervention preferences provide strong correction-forecasting performance. Combining both sources of information performs strongly across both tasks, even when generalizing to participants unseen during training. These findings demonstrate that both when and how humans correct provide complementary information for learning value representations from corrective interactions.
\end{abstract}




\maketitle

\pagestyle{plain}

\begin{figure}
 \includegraphics[width=\linewidth]{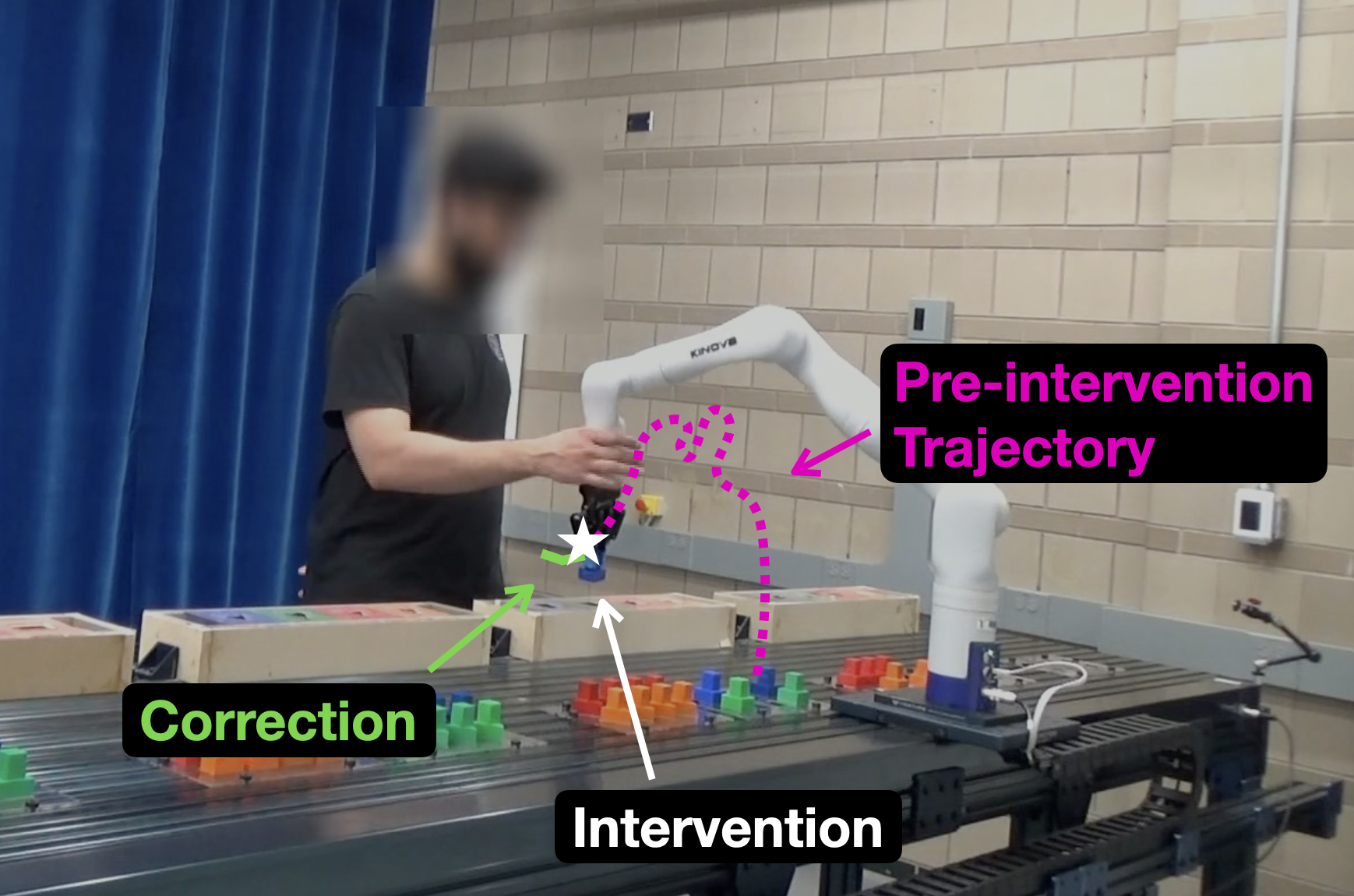}
     \caption{Components of a human corrective interaction: the robot's pre-intervention trajectory, the human's intervention, and the subsequent correction. 
     }
  \label{fig:first}
\end{figure}

\begin{figure*}
 \includegraphics[width=\textwidth]{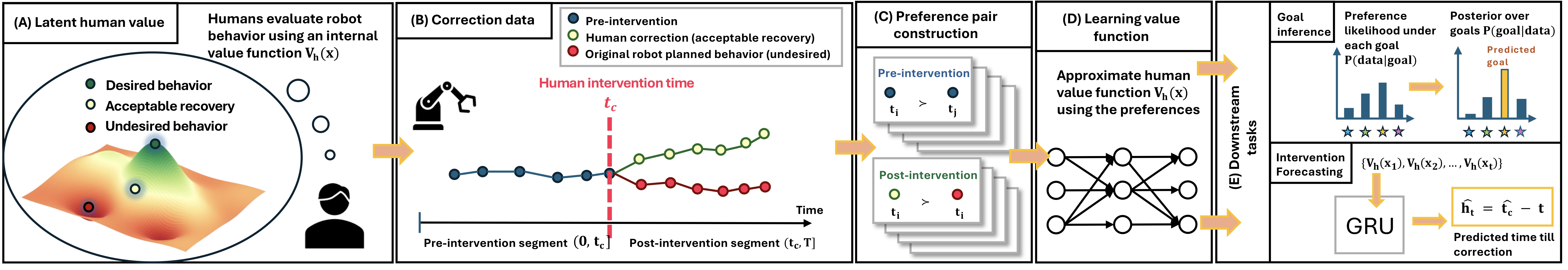}
    \caption{Overview of our approach. The human has an internal, latent value function $V_h(x)$ that defines acceptable, undesirable, and desirable behaviors \textbf{(A)}. This is manifested in their decisions to allow, interrupt, and modify the robot’s behaviors, respectively. When they correct the robot \textbf{(B)}, these decisions are recorded as (i) the robot’s originally-planned trajectory, (ii) the timestep at which the human intervened, and (iii) the corrected trajectory.
We then convert this correction data into sets of pre- and post-intervention pairwise preferences \textbf{(C)} for training an approximation of the human’s value function \textbf{(D)}.
Since we cannot directly observe the human’s latent value function, we define two downstream tasks \textbf{(E)} to assess whether the learned value function produces the same observable outcomes as real users’ corrections. }
  \label{fig:teaser}
\end{figure*}
	
\section{Introduction}

Human corrections provide a natural way for people to communicate how a robot should behave. When a person observes undesirable robot behavior, they can intervene and redirect the robot toward a more acceptable behavior. These interactions provide rich feedback about the person's preferences without requiring them to explicitly specify a reward function or provide complete demonstrations \cite{losey2022physical, losey2018including, jin2022learning, bobu2022inducing, wang2024toward}.
If a robot can learn the human's underlying value function from corrections, it can optimize its future behaviors accordingly, reducing the likelihood that the human will need to correct it again in the future~\cite{losey2022physical}. 

Learning such a value function from corrections, however, remains challenging. Corrections only reveal how the person would modify the robot's behavior from a particular state, rather than directly specifying how they evaluate all possible behaviors \cite{cui2021understanding}. Furthermore, they are prompted by the human's subjective decision to intervene in the robot's behavior, which does not always align with the robot's actual success or failure to complete a task~\cite{wang2025effects}. Prior work has established how a robot can learn from corrections by treating them as demonstrations of desired behavior \cite{zhang2019learning, bajcsy2017learning, bajcsy2018learning, gribovskaya2011motion} or as supervision for adapting its policy \cite{losey2022physical, korkmaz2025mile}. These approaches primarily leverage the correction itself (shown in green in Fig~ \ref{fig:first}), while the trajectory leading up to the person's decision to intervene remains comparatively underexplored (shown in pink in Fig. \ref{fig:first}).


In this paper, we investigate how to translate the temporal context surrounding a correction into pairwise preferences between robot behaviors, with the goal of learning the human's underlying value function. We explore multiple hypotheses for how a human's value function guides their decision to intervene in, and subsequently modify, the robot's behavior. We evaluate these hypotheses on a set of 876 corrections provided by real users, and discover these relative relationships provide effective supervision for learning value representations from human corrections.

 We evaluate the learned value functions through goal inference and correction forecasting. Our results show that preferences constructed from post-intervention correction data are particularly effective for goal inference, while preferences constructed from pre-intervention trajectories provide strong correction forecasting performance. 
 Combining both pre- and post-intervention data results in strong performance across both evaluation tasks, while also generalizing to participants unseen during training. Together, these results suggest that learning from corrections does not require treating human feedback as optimal behavior; the local, relative judgments revealed through corrective interactions can instead provide sufficient structure for learning useful value representations.
Our contributions are as follows:
\begin{enumerate}
    \item We formulate Learning from Corrections as a problem of constructing and learning from \emph{pairwise preferences} derived from both pre- and post-intervention trajectories.
    \item We propose two new methods for constructing pre-intervention preference pairs based on how the perceived value of the robot's performance changes throughout its trajectory.
    \item Using a large dataset of corrections from real users (876 datapoints from N=120 users, derived from~\cite{wang2025effects}), we evaluate four techniques for constructing pairwise preferences from pre- and post-intervention trajectories, assessing how well they reflect users' value functions to support both goal inference and correction forecasting.
\end{enumerate}

\section{Related Work}
\label{sec:related_work}

\paragraph{Learning from Corrections and Demonstrations.} A substantial body of prior work learns from human corrections by treating corrected robot behavior as demonstrations or informative human actions. For example, \citet{zhang2019learning} formulate local corrections as partial demonstrations and generalize them to infer the user's intended trajectory for objective learning. Other approaches \cite{bajcsy2017learning, losey2022physical} model corrections as purposeful actions that provide information about latent task parameters and use them to adapt the robot's objective. However, the behavior demonstrated during a correction need not represent the human's globally desired behavior. Instead, it may represent an acceptable recovery that is preferred to the robot's behavior at the point of intervention. Moreover, corrections can be difficult to interpret when the robot's hypothesis space does not adequately represent the human's preferences; \citet{bobu2020quantifying}, for example, estimate confidence in whether an observed correction can be explained by the robot's current model. These works demonstrate that corrective actions provide rich information about human objectives, but primarily focus on interpreting the action taken after the human decides to intervene.

\paragraph{Preference Learning from Human Feedback.} Corrections can also provide relative information: rather than assuming that corrected behavior is globally optimal, the correction can indicate that one behavior is preferred to another. Prior work has used comparison-based feedback to infer human objectives and improve robot behavior \cite{jain2013learning, jain2015learning, wilde2020learning, losey2018including}. Corrections have similarly been interpreted as evidence that corrected behavior is locally preferred to alternative robot behaviors \cite{10.1145/3623384}. Such relative comparisons are useful when the human's latent reward is not directly observable and the provided behavior need not constitute an optimal demonstration. Preference learning has also been used to learn richer reward representations; for example, \citet{biyik2020active} consider nonlinear reward learning when linear reward representations are insufficient to capture human preferences. Our work builds on this preference-based perspective by constructing preference relationships directly from corrections and using them to learn nonlinear value functions that approximate the human's latent evaluation of robot behavior.

\paragraph{Learning from Intervention Timing.} Corrective interactions provide information not only through \emph{how} a human corrects, but also through \emph{when} they decide to intervene. Prior work has shown that intervention and non-intervention decisions can provide supervision about robot behavior. \citet{spencer2022expert} use both interventions and non-interventions to constrain robot value estimates during online learning, while \citet{korkmaz2025mile} models intervention decisions based on the relationship between human and robot value estimates. \citet{wang2026enhancing} further demonstrates that correction timing provides information about human evaluations of robot behavior and can improve inference of the human's intended task objective. These approaches establish intervention timing as an informative signal, but do not explicitly utilize the temporal progression of behavior leading to an intervention for learning the human's value function.

\paragraph{Research Gap.}
Prior work has largely focused on learning from the post-intervention correction trajectory, thus underutilizing its context provided by the pre-intervention trajectory. Moreover, although corrections inherently reveal relative relationships between robot behaviors, these relationships have not been explicitly formulated as pairwise preferences for learning value functions. We address these gaps by constructing preference relationships from both pre- and post-intervention behavior.

\section{Learning Value Functions from Corrections}
\label{sec:approach}



Our overall approach (Fig.~\ref{fig:teaser}) is to formulate Learning from Corrections as a problem of constructing and learning from pairwise preferences. This allows us to leverage extensive prior work on inverse reinforcement learning from preferences \cite{furnkranz2012preference,wirth2017survey,christiano2017deep,biyik2018batch,biyik2020active}. The primary challenge we address involves \emph{how} to derive pairwise preferences from correction data.

We extract preference information from both sides of the intervention (Fig.~\ref{fig:teaser}B--C). \emph{Post-intervention preferences} capture \emph{how} the human corrects by comparing the corrected behavior with the robot behavior it replaces. \emph{Pre-intervention preferences} capture \emph{when} the human corrects by constructing relative comparisons among robot behaviors leading up to the intervention.
We use these preferences to learn value functions that approximate the human's latent evaluation of robot behavior (Fig.~\ref{fig:teaser}D). We investigate alternative hypotheses for constructing pre-intervention preferences, learn from pre- and post-intervention information independently and in combination, and evaluate the resulting representations through goal inference and correction forecasting (Fig.~\ref{fig:teaser}E).

\subsection{Formulating LfC as Preference-Based Value Learning}

A human supervises a robot executing a trajectory $\xi=[\langle s_t,a_t\rangle]_{t=1}^{T}$ consisting of state-action pairs. At any timestep, the human may intervene and pause the robot's execution. We denote the intervention timestep by $t_c$, which partitions the robot's originally planned trajectory into the executed pre-intervention segment $\xi^{pre}=[\langle s_t,a_t\rangle]_{t=1}^{t_c}$ and the unexecuted planned continuation $\xi^{post}=[\langle s_t,a_t\rangle]_{t=t_c}^{T}$. After intervening, the human provides a corrective trajectory $\xi^{corr}$ beginning from the intervention state $s_{t_c}$.

We represent the human's latent evaluation of robot behavior using a learned value function $V(x^g)$, where $x^g=\phi(\xi,g)$ or $x_t^g=\phi(s_t,g)$ denotes a feature representation of trajectory $\xi$ or state $s_t$ with respect to a candidate task goal $g$. Given a preference $x_a^g \succ x_b^g$, we seek to learn $V$ such that $V(x_a^g)>V(x_b^g)$, without requiring either behavior to be optimal. We adopt a standard preference-learning formulation for learning $V$ from such pairwise comparisons~\cite{wirth2017survey, christiano2017deep, biyik2018batch}.

Given a preference dataset $\mathcal{D}_g=\{(x_a^g,x_b^g,y)\}$ for candidate goal $g$, where $y=+1$ indicates $x_a^g \succ x_b^g$ and $y=-1$ indicates $x_b^g \succ x_a^g$, we model the likelihood of the observed preferences as
\begin{equation}
\label{eq:preflikelihood}
\log P(\mathcal{D}_g \mid g)
=
\sum_{(x_a^g,x_b^g,y)\in\mathcal{D}_g}
\log
\Theta\left(
\frac{
y\left[V(x_a^g)-V(x_b^g)\right]
}{
\sigma_p
}
\right),
\end{equation}
where $\Theta(\cdot)$ denotes the standard normal cumulative distribution function and $\sigma_p$ controls the assumed noise in the observed preferences. Our primary contribution lies in how $\mathcal{D}_g$ is constructed from human corrections that lead to better value learning and downstream task performance. The following sections introduce and evaluate alternative preference relationships derived from behavior before and after intervention.

\subsection{Post-Intervention Preferences: How Humans Correct}

Prior work on Learning from Corrections commonly assumes that the behavior supplied by the human after intervention is preferable to the robot behavior it replaces, using corrections to update the robot's objective or behavior toward the human-provided alternative~\cite{zhang2019learning,bajcsy2017learning,bajcsy2018learning,losey2022physical}. We express this established assumption as a trajectory-level pairwise preference between the corrected trajectory and the robot's planned continuation. Importantly, this relationship only establishes a relative preference: the corrected behavior represents a more acceptable alternative, but is not assumed to be the human's globally desired or optimal behavior. We additionally consider a finer-grained timestep formulation that constructs preferences between behaviors at corresponding timesteps within the two trajectories.

\paragraph{Trajectory-Level Preferences.} For a trajectory segment $\xi$, we define its value as the average of its timestep values,
\begin{equation}
V(x^g)=V(\phi(\xi,g))
=
\frac{1}{|\xi|}
\sum_{t\in\xi}
V\!\left(\phi(s_t,g)\right).
\end{equation}
The corrected trajectory is preferred to the robot's original planned continuation; i.e., 
$V(x^{corr,g}) > V(x^{post,g})$.
This formulation represents the correction as a single trajectory-level preference.

\paragraph{Timestep-Level Preferences.} A trajectory-level comparison may obscure localized differences between the robot's planned behavior and the human's correction. We therefore also construct timestep-level preferences. Because $\xi^{post}$ and $\xi^{corr}$ may have different durations, both trajectories are resampled onto a common normalized progress grid of length $M=30$. At each aligned progress index $j$,
\begin{equation}
V(\tilde{x}^{corr,g}_j)
>
V(\tilde{x}^{post,g}_j),
\qquad j=1,\ldots,M.
\end{equation}
These comparisons provide localized preference supervision between the corrected behavior and the behavior it replaces.

\subsection{Pre-Intervention Temporal Preferences: When Humans Correct}

We pose two hypotheses for translating the temporal progression of behavior preceding an intervention into preference pairs.

\paragraph{Monotonic Decreasing Hypothesis.} Our monotonic decreasing formulation is motivated by continuous action monitoring \cite{vidal2020errors} suggesting that humans continuously evaluate behavior and intervene once accumulated deviations become sufficiently large. We therefore hypothesize that the value of the robot's behavior decreases monotonically as intervention approaches. 
We consider two strategies for translating this hypothesis into preference pairs. The \emph{neighboring} formulation compares consecutive timesteps,
\begin{equation}
V(x_t^g)>V(x_{t+1}^g) \qquad \forall t<t_c
\end{equation}
whereas the \emph{all-pairs} formulation compares any pair of temporally ordered timesteps,
\begin{equation}
V(x_i^g)>V(x_j^g)
\qquad \forall i, j \;\; | \;\; i<j<t_c.
\end{equation}
Because the number of all-pairs comparisons grows quadratically with trajectory length, we randomly subsample preference pairs during training.

\paragraph{Peak-and-Drop Hypothesis.} A monotonic decrease may not capture cases in which the robot initially appears to make progress before its behavior deteriorates. Motivated by the peak-end rule \cite{wang2024continuous} suggesting that human evaluations emphasize salient moments within an interaction, we propose that perceived value reaches a local peak before decreasing until the intervention. We assume a latent peak timestep $t_p<t_c$ such that:
\begin{align}
    V(x_{t+1}^g)&>V(x_t^g) \quad && \forall t<t_p\\
V(x_t^g)&>V(x_{t+1}^g) && \forall t \;\;|\;\; t_p\leq t<t_c.
\end{align}
As with the monotonic-decreasing hypothesis, we consider both neighboring and all-pairs preference constructions.

Because $t_p$ is not observable, we infer it jointly with the value function. Candidate peaks are restricted to a temporal window preceding intervention and must satisfy a minimum post-peak drop criterion. We use an Expectation Maximization-style procedure that alternates between estimating the most likely peak location for each trajectory and optimizing the value function under the resulting preference pairs until convergence (details in Appendix~\ref{sec: em alg}).

\subsection{Combining Pre- and Post-Intervention Preferences}
\label{sec:temporal_two_value}

\paragraph{Temporal Two-Value.} The pre- and post-intervention formulations provide different preference supervision and may capture complementary aspects of the human's evaluation. We therefore combine the strongest individual formulations from each source while preserving their learned value representations separately.

We learn two independent value functions,
\begin{equation}
V_{\mathrm{when}}(x^g)=f_{\theta_{\mathrm{when}}}(x^g),
\qquad
V_{\mathrm{how}}(x^g)=f_{\theta_{\mathrm{how}}}(x^g).
\end{equation}
Here, $V_{\mathrm{when}}$ is learned from pre-intervention preferences, while $V_{\mathrm{how}}$ is learned from corrected-over-planned post-intervention preferences. Both are optimized independently using Eq.~\ref{eq:preflikelihood}, and their representations are combined only during downstream inference.

\section{Evaluation}
\label{sec:evaluation}

\subsection{MILE-Based Value Learning Baseline}
\label{sec:mile}

We compare our preference-based formulations with MILE~\cite{korkmaz2025mile}, an existing approach that learns from both intervention decisions and corrective actions. MILE distinguishes between a human-preferred policy and a mental model of the robot. Because our objective is to learn value functions rather than policies, we adapt this structure to two value functions, which we refer to as \textsc{MILE-2V}.

For each candidate goal $g$, we represent robot behavior using the same goal-conditioned features as our preference-based methods, $x_t^g=\phi(s_t,g)$. We learn two independent nonlinear value functions,
\begin{equation}
V_h(x_t^g)=f_{\theta_h}(x_t^g),
\qquad
V_r(x_t^g)=f_{\theta_r}(x_t^g),
\end{equation}
where $V_h$ represents the human-preferred value and $V_r$ is the value-based analogue of MILE's mental model of the robot's behavior. Both use the same nonlinear architecture as our other nonlinear value models but have independent parameters.

\paragraph{Learning when the human intervenes.} For each corrected trajectory, we assign $\nu_t=0$ for timesteps before intervention and $\nu_{t_c}=1$ at the intervention timestep $t_c$. Following MILE, we assume that intervention becomes more likely when the human-preferred value exceeds the robot's expected value by more than an intervention threshold $c$. Because we learn values rather than policies, we approximate the robot's expected value by sampling candidate successor states. For each observed transition $x_{t-1}^g\rightarrow x_t^g$, we construct a set $\mathcal{X}_t=\{\hat{x}_t^{(1)},\ldots,\hat{x}_t^{(K)}\}$ containing the observed successor and $K-1$ alternatives sampled at the same transition distance from $x_{t-1}^g$. We induce a Boltzmann distribution over these candidates using $V_r$,
\begin{equation}
\pi_r(\hat{x}_t^{(k)}\mid x_{t-1}^g)
=
\frac{\exp(V_r(\hat{x}_t^{(k)})/\tau_r)}
{\sum_{\ell=1}^{K}\exp(V_r(\hat{x}_t^{(\ell)})/\tau_r)}.
\end{equation}
We then approximate the robot's expected value as
\begin{equation}
\bar{V}_r(x_t^g)
=
\sum_{k=1}^{K}
\pi_r(\hat{x}_t^{(k)}\mid x_{t-1}^g)
V_r(\hat{x}_t^{(k)}).
\end{equation}
The intervention probability is represented as
\begin{equation}
\hat{\nu}_t^g
=
\Phi\left(
\frac{V_h(x_t^g)-\bar{V}_r(x_t^g)-c}{\sigma_m}
\right),
\label{eq:mile_value_intervention}
\end{equation}
where $\Phi(\cdot)$ is the standard normal cumulative distribution function and $\sigma_m$ controls intervention noise. Because each trajectory contains many non-intervention timesteps but only one intervention timestep, we use a class-balanced binary cross-entropy loss,
\begin{equation}
\mathcal{L}_{\mathrm{when}}^{\mathrm{MILE}}
=
-\frac{1}{N_{\mathrm{when}}}
\sum_{i,t}
\left[
w_1\nu_{i,t}\log\hat{\nu}_{i,t}^g
+
w_0(1-\nu_{i,t})\log(1-\hat{\nu}_{i,t}^g)
\right].
\label{eq:mile_when_loss}
\end{equation}
This loss jointly supervises $V_h$, $V_r$, and the intervention threshold $c$.

\paragraph{Learning how the human intervenes.} Rather than representing the correction as a pairwise preference between corrected and planned behavior, \textsc{MILE-2V} models the next state selected by the human during the correction. We resample each corrective trajectory onto a normalized progress grid of length $M=30$ and consider each transition $x_t^{\mathrm{corr},g}\rightarrow x_{t+1}^{\mathrm{corr},g}$. For each transition, we construct a candidate successor set containing the observed successor and $K-1$ alternatives sampled at the same transition distance. Using the human-preferred value function, we define
\begin{equation}
P(\hat{x}_{t+1}^{(k)}\mid x_t^{\mathrm{corr},g})
=
\frac{\exp(V_h(\hat{x}_{t+1}^{(k)})/\tau_h)}
{\sum_{\ell=1}^{K}\exp(V_h(\hat{x}_{t+1}^{(\ell)})/\tau_h)}.
\end{equation}
We train $V_h$ by maximizing the likelihood of the successor selected by the human,
\begin{equation}
\mathcal{L}_{\mathrm{how}}^{\mathrm{MILE}}
=
-\frac{1}{N_{\mathrm{how}}}
\sum_{i,t}
\log P(x_{i,t+1}^{\mathrm{corr},g}\mid x_{i,t}^{\mathrm{corr},g}).
\label{eq:mile_how_loss}
\end{equation}
Unlike the intervention loss, which jointly supervises $V_h$ and $V_r$, the corrective-transition loss directly supervises only $V_h$. This mirrors MILE, where intervention decisions update both the human-preferred model and the robot mental model, while corrective actions update only the human-preferred model.

\subsection{Evaluation Tasks}
\label{sec:evaluation_tasks}
Because the human's latent value function is unobservable, we cannot directly compare the learned value functions against ground-truth human values. Instead, we evaluate whether the learned representations capture properties of human preferences through two downstream proxy tasks: goal inference and correction forecasting.

\paragraph{Goal Inference.} We evaluate whether the learned value representations capture information about the human's intended task objective. Candidate goals are discretized over a 2D workspace grid with $x\in[-0.2,0.2]$ m and $y\in[-0.3,0.3]$ m at $0.01$ m resolution. For each candidate goal $g_j$, the observed correction is represented relative to that goal and assigned a goal-conditioned score. For the single-value preference models, this score is the average preference log-likelihood,
\begin{equation}
S_{\mathrm{pref}}(g_j)
=
\frac{1}{|\mathcal{D}_{g_j}|}
\log P(\mathcal{D}_{g_j}\mid g_j).
\label{eq:pref_goal_score}
\end{equation}

For \textsc{Temporal Two-Value}, we compute separate average preference log-likelihood scores from the pre- and post-intervention value functions, denoted $S_{\mathrm{when}}(g_j)$ and $S_{\mathrm{how}}(g_j)$, respectively, and combine them with equal weighting,
\begin{equation}
S_{\mathrm{2V}}(g_j)
=
\frac{1}{2}
\left[
S_{\mathrm{when}}(g_j)
+
S_{\mathrm{how}}(g_j)
\right].
\label{eq:two_value_goal_score}
\end{equation}

For \textsc{MILE-2V}, we similarly compute separate goal-conditioned scores from its intervention and corrective-transition models. Let $S_{\mathrm{when}}^{\mathrm{MILE}}(g_j)$ denote the average class-balanced intervention log-likelihood corresponding to Eq.~\ref{eq:mile_when_loss}, and let $S_{\mathrm{how}}^{\mathrm{MILE}}(g_j)$ denote the average corrective-transition log-likelihood corresponding to Eq.~\ref{eq:mile_how_loss}. We combine these scores with equal weighting,
\begin{equation}
S_{\mathrm{MILE}}(g_j)
=
\frac{1}{2}
\left[
S_{\mathrm{when}}^{\mathrm{MILE}}(g_j)
+
S_{\mathrm{how}}^{\mathrm{MILE}}(g_j)
\right].
\label{eq:mile_goal_score}
\end{equation}

For each model, we normalize the corresponding goal-conditioned score $S(g_j)$ across candidate goals to obtain
\begin{equation}
p(g_j\mid\mathcal{D})
=
\frac{\exp(S(g_j))}
{\sum_{g_k\in\mathcal{G}}\exp(S(g_k))}.
\label{eq:goal_posterior}
\end{equation}

During training, the inferred goal distribution is supervised using a Gaussian target centered at the ground-truth task goal $g^*$,
\begin{equation}
q(g_j)
=
\frac{
\exp\left(
-\|g_j-g^*\|^2 / (2\sigma_g^2)
\right)
}{
\sum_{g_k\in\mathcal{G}}
\exp\left(
-\|g_k-g^*\|^2 / (2\sigma_g^2)
\right)
},
\label{eq:gaussian}
\end{equation}
where $\sigma_g=0.03$. Performance is measured using $D_{\mathrm{KL}}(q(g)\|p(g\mid\mathcal{D}))$ on held-out trajectories, with lower values indicating greater agreement with the target distribution.

\paragraph{Correction Forecasting.} We next evaluate whether the learned value representations contain temporal information predictive of an upcoming intervention. At each pre-intervention timestep $t$, we evaluate the learned representation at the ground-truth goal $g^*$ and predict the remaining intervention horizon $h_t=t_c-t$ from its recent history. For models that learn a single value function, we construct
\begin{equation}
H_t=[v_{t-L+1},\ldots,v_t],
\qquad
v_s=V(x_s^{g^*}),
\label{eq:forecast_history}
\end{equation}
with history length $L=100$. For \textsc{MILE-2V}, forecasting uses $V_h$. For \textsc{Temporal Two-Value}, each timestep retains both learned representations,
\begin{equation}
H_t=[\mathbf{v}_{t-L+1},\ldots,\mathbf{v}_t], \qquad \mathbf{v}_s
=
\left[
V_{\mathrm{when}}(x_s^{g^*}),
V_{\mathrm{how}}(x_s^{g^*})
\right].
\end{equation}


We apply temporal weighting to the value history so that observations closer to the current timestep exert greater influence. For a history of length $L$, the weighted representation at timestep $t$ is
\begin{equation}
\tilde{H}_t
=
[\gamma^{L-1}v_{t-L+1},\ldots,\gamma v_{t-1},v_t],
\label{eq:temporal_weighting}
\end{equation}
where $\gamma\in(0,1]$ is the temporal decay factor. Smaller $\gamma$ increasingly emphasizes recent values, while $\gamma=1$ preserves the value history without temporal decay.

To emphasize observations closer to the intervention, which are more directly informative about the upcoming correction, we apply a Gaussian weighting based on the remaining intervention horizon $h_t=t_c-t$,
\begin{equation}
w(h_t)
=
\exp\left(
-h_t^2 / (2\sigma_h^2)
\right),
\label{eq:horizon_weighting}
\end{equation}
where $\sigma_h$ controls how broadly this emphasis extends over time. This weighting reduces the influence of temporally distant observations while retaining information from a range of pre-intervention behavior, preventing training from being dominated by samples far from the correction. The forecasting objective is then
\begin{equation}
\mathcal{L}_{\mathrm{forecast}}
=
\frac{
\sum_t w(h_t)\,
\ell_{\mathrm{SL1}}(\hat{h}_t,h_t)
}{
\sum_t w(h_t)
},
\label{eq:forecast_loss}
\end{equation}
where $\ell_{\mathrm{SL1}}$ denotes the Smooth L1 loss. Smaller $\sigma_h$ focuses training more strongly on behavior immediately preceding intervention, whereas larger $\sigma_h$ incorporates information over a broader range of intervention horizons.

Forecasting performance is evaluated using mean absolute error (MAE) between the predicted and ground-truth intervention horizons in the held-out set,
\begin{equation}
\mathrm{MAE}
=
\frac{1}{N}
\sum_{i=1}^{N}
|\hat{h}_i-h_i|,
\end{equation}
with results reported in seconds using the dataset timestep duration of $0.1\,\mathrm{s}$. Lower MAE indicates that the predicted intervention horizon is closer to the true time remaining until intervention.

\subsection{Experimental Setup}
\label{sec:experimental_setup}

We evaluate our work using the correction data collected by~\citet{wang2025effects}, in which $N=120$ participants supervised a 7-DoF Kinova Gen3 robotic arm during pick-and-place trials and could intervene during execution to provide corrective feedback. 
From their dataset, we use corrections occurring within the first $90\%$ of the original trajectory duration to ensure sufficient pre- and post-intervention behavior to construct the corresponding preference pairs, which leads to 876 corrected trajectories. These trajectories are randomly divided into $60\%$ training, $10\%$ validation, and $30\%$ test sets. 

Robot behavior is represented using goal-conditioned features $x_t^g=\phi(s_t,g)$ capturing motion alignment, consistency, legibility, and task progress (details in Appendix~\ref{sec: features}). Feature vectors are standardized using statistics computed from the training set. We evaluate four individual preference formulations: \textsc{Trajectory} and \textsc{Timestep} for post-intervention preferences, and \textsc{Decreasing} and \textsc{Peak-and-Drop} for pre-intervention preferences. We additionally evaluate \textsc{Temporal Two-Value} and the \textsc{MILE-2V} baseline.

For model selection, we evaluate both neighboring and all-pairs variants of the pre-intervention formulations and compare linear and nonlinear value functions where applicable. The linear value function is $V_{\mathrm{lin}}(x^g)=w^\top x^g$. Nonlinear value functions use a multilayer perceptron with two hidden layers of 64 units and ReLU activations followed by a scalar output. Preference likelihoods use $\sigma_p=0.1$. For correction forecasting, value histories are provided to a two-layer GRU with hidden dimension 128 and dropout $0.1$, followed by a scalar output predicting the remaining intervention horizon.

\subsection{Evaluation Protocol}
\label{sec:evaluation_protocol}

Our evaluation proceeds in stages: (1) model and hyperparameter selection, (2) comparison of the final six models, and (3) additional analyses of participant generalization and temporal randomization.

\paragraph{Temporal-Range Evaluation.} For model and hyperparameter selection, we divide corrections according to when they occur within the original trajectory into three non-overlapping ranges: $0$--$70\%$, $70$--$80\%$, and $80$--$90\%$ of trajectory progress. Models are trained and evaluated separately within each range, with 10 independent runs per range. We aggregate performance across the three ranges, yielding 30 runs for each configuration. This stratification reduces sensitivity to any single intervention range when selecting model structure and hyperparameters.

\paragraph{Model and Hyperparameter Selection.} We compare neighboring and all-pairs pre-intervention preferences and linear and nonlinear value functions. For correction forecasting, we additionally evaluate temporal decay factor $\gamma\in\{0.7,0.8,0.9,1.0\}$ and Gaussian weighting parameter $\sigma_h\in\{15,25,50\}$, aggregating results for each $\sigma_h$ across $\gamma$'s and for each $\gamma$ across $\sigma_h$'s. Based on these experiments, we select the preference construction, value architecture, and forecasting hyperparameters used in subsequent comparisons.

\paragraph{Comparison of Preference Formulations.} We compare the six representative models 
using the selected value function architecture and preference construction for the pre-intervention models. The same selected forecasting configuration is used for all models.

\paragraph{Randomization Control.} Additionally, we test whether correction forecasting depends on the temporal ordering of the learned values. This validates that our models reflect how interventions are prompted by temporal patterns in the value of the robot's behavior,
rather than forecasting relying only on the distribution or magnitude of values. Using the temporal-range protocol, we compare forecasting from the original value histories with histories containing the same values in shuffled timestep order while holding the forecasting architecture, data splits, and hyperparameters fixed.

\paragraph{Generalization Across Participants.} Finally, we evaluate the six models on corrections from unseen participants. For this analysis, all corrections occurring within the first $90\%$ of trajectory progress are pooled before training, ensuring that each correction contains sufficient pre- and post-intervention behavior for constructing and evaluating both sources of preference information. Each model is evaluated over 30 independent runs. Under the \emph{trajectory-level split}, trajectories are partitioned without enforcing participant separation, such that trajectories from the same participant may appear across partitions. Under the \emph{participant-level split}, all trajectories from each participant are assigned exclusively to one partition, ensuring that test participants are unseen during training. The same final model configurations are used under both splits.

\section{Results}
\label{sec:results}

\begin{table}[t]
\centering
\caption{Goal-inference performance across value architectures and preference constructions. We report mean KL divergence 
$\pm$ 95\% confidence interval. Lower values are better.}
\label{tab:goal_model_selection}
\small
\begin{tabular}{lcc}
\toprule
& \textbf{Linear} & \textbf{Nonlinear} \\
\midrule
\textbf{Trajectory}
& $1.704 \pm 0.181$
& $0.703 \pm 0.354$ \\

\textbf{Timestep}
& $1.574 \pm 0.111$
& $0.073 \pm 0.046$ \\

\textbf{Peak}
& $2.570 \pm 0.115$
& $0.239 \pm 0.064$ \\

\textbf{Peak All-Pairs}
& $2.884 \pm 0.024$
& $0.500 \pm 0.092$ \\

\textbf{Decreasing}
& $2.095 \pm 0.183$
& $0.159 \pm 0.027$ \\

\textbf{Decreasing All-Pairs}
& $2.043 \pm 0.183$
& $0.216 \pm 0.080$ \\
\bottomrule
\end{tabular}
\end{table}

\begin{table*}[t]
\centering
\caption{Correction-forecasting performance across Gaussian weighting parameters $\sigma_h$ (left) and temporal decay factors $\gamma$ (right) for the nonlinear value models. Values report mean MAE in seconds $\pm$ 95\% confidence interval (lower is better). 
}
\label{tab:hyperparameter_selection}
\small
\begin{tabular}{lccc|cccc}
\toprule
& \multicolumn{3}{c|}{\textbf{Gaussian Weighting ($\sigma_h$)}}
& \multicolumn{4}{c}{\textbf{Temporal Decay ($\gamma$)}} \\
\cmidrule(lr){2-4} \cmidrule(lr){5-8}
& $\boldsymbol{\sigma_h = 15}$ & $\boldsymbol{\sigma_h = 25}$ & $\boldsymbol{\sigma_h = 50}$
& $\boldsymbol{\gamma = 0.7}$ & $\boldsymbol{\gamma = 0.8}$ & $\boldsymbol{\gamma = 0.9}$ & $\boldsymbol{\gamma = 1.0}$ \\
\midrule
\textbf{Trajectory}
& $2.030 \pm 0.071$
& $1.356 \pm 0.045$
& $\boldsymbol{1.119 \pm 0.035}$
& $1.683 \pm 0.101$
& $1.531 \pm 0.093$
& $1.440 \pm 0.092$
& $\boldsymbol{1.353 \pm 0.101}$ \\

\textbf{Timestep}
& $2.169 \pm 0.070$
& $1.448 \pm 0.044$
& $\boldsymbol{1.182 \pm 0.038}$
& $1.746 \pm 0.106$
& $1.604 \pm 0.108$
& $\boldsymbol{1.498 \pm 0.095}$
& $1.550 \pm 0.107$ \\

\textbf{Peak}
& $2.166 \pm 0.045$
& $1.390 \pm 0.037$
& $\boldsymbol{1.109 \pm 0.027}$
& $1.740 \pm 0.082$
& $1.614 \pm 0.086$
& $1.462 \pm 0.075$
& $\boldsymbol{1.403 \pm 0.076}$ \\

\textbf{Peak All-Pairs}
& $2.162 \pm 0.067$
& $1.461 \pm 0.050$
& $\boldsymbol{1.204 \pm 0.039}$
& $1.776 \pm 0.120$
& $1.641 \pm 0.120$
& $1.516 \pm 0.108$
& $\boldsymbol{1.503 \pm 0.110}$ \\

\textbf{Decreasing}
& $1.890 \pm 0.049$
& $1.161 \pm 0.033$
& $\boldsymbol{0.938 \pm 0.027}$
& $1.485 \pm 0.068$
& $1.412 \pm 0.073$
& $1.249 \pm 0.069$
& $\boldsymbol{1.175 \pm 0.077}$ \\

\textbf{Decreasing All-Pairs}
& $2.366 \pm 0.096$
& $1.443 \pm 0.054$
& $\boldsymbol{1.140 \pm 0.033}$
& $1.952 \pm 0.149$
& $1.778 \pm 0.153$
& $1.571 \pm 0.132$
& $\boldsymbol{1.434 \pm 0.101}$ \\
\bottomrule
\end{tabular}
\end{table*}

\subsection{Model and Hyperparameter Selection}
\label{sec:model_selection}

We first report the model and parameter selection results described in Sec.~\ref{sec:evaluation_protocol}, aggregated across 30 runs from the three intervention ranges. Results are presented in Tables 1-2, with corresponding plots and pairwise significance annotations provided in Appendix~\ref{app:significance_plots}.

\paragraph{Value architecture and preference construction.} Table~\ref{tab:goal_model_selection} compares goal-inference performance across value architectures and preference constructions. Nonlinear value functions substantially reduce KL divergence across the preference formulations. Within the nonlinear models, Timestep and Decreasing achieve particularly low KL divergence. For both Peak-and-Drop and Decreasing, neighboring preferences yield lower KL divergence than all-pairs preferences. We therefore use nonlinear value functions and neighboring pre-intervention preferences in subsequent experiments.

\paragraph{Correction-forecasting parameters.} Table~\ref{tab:hyperparameter_selection}, left, compares forecasting performance across $\sigma_h\in\{15,25,50\}$. Forecasting error generally decreases as $\sigma_h$ increases, with $\sigma_h=50$ yielding the lowest MAE across the evaluated formulations. 
We additionally evaluate $\gamma\in\{0.7,0.8,0.9,1.0\}$ (Table~\ref{tab:hyperparameter_selection}, right). Although performance varies across formulations, $\gamma=1.0$ yields the lowest or among the lowest MAE for most models. We therefore use $\gamma=1.0$ and $\sigma_h=50$ for all subsequent correction-forecasting experiments.



\begin{figure}
    \centering
    \includegraphics[width=\linewidth]{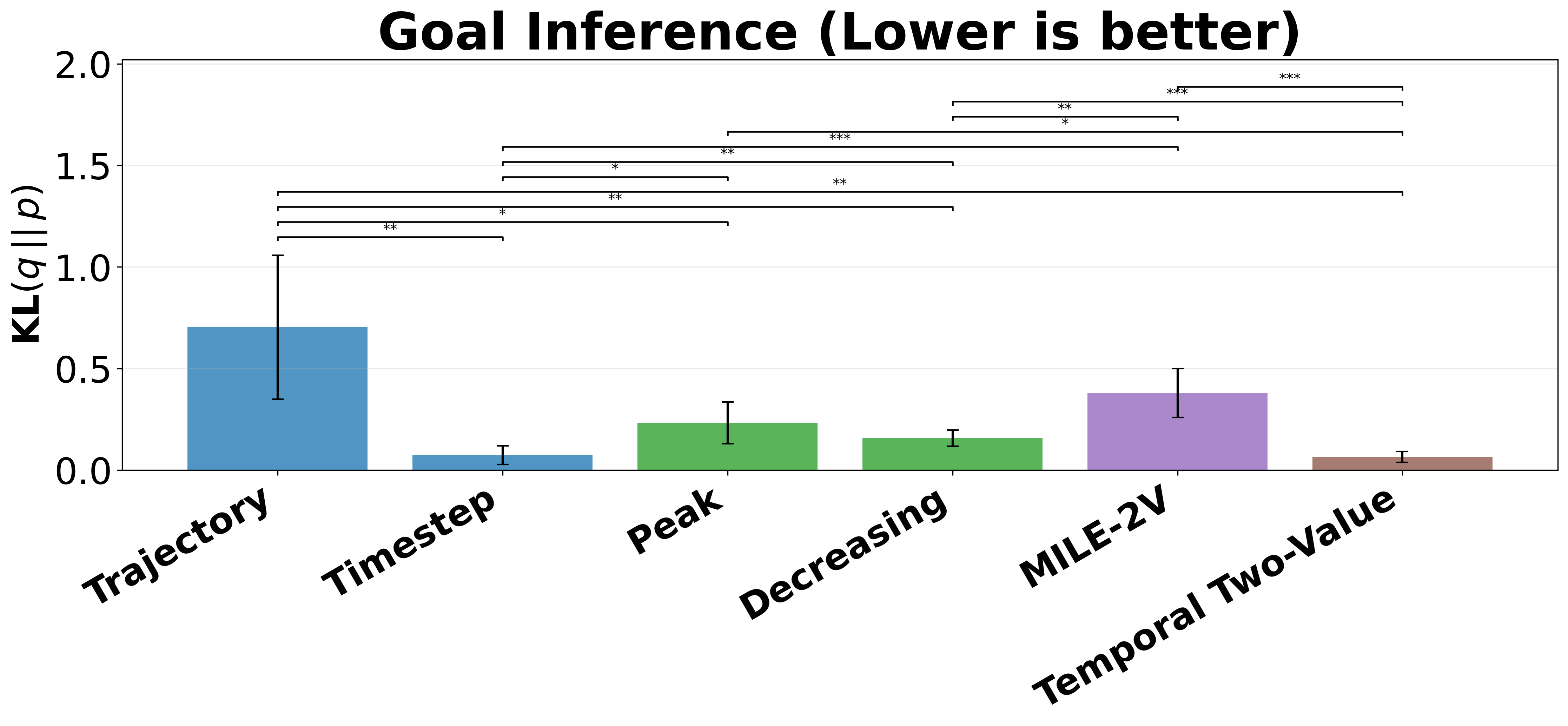}
    \caption{Goal-inference performance across the six representative models. Bars show mean KLD with 95\% CI. Brackets indicate significance ($^*p<0.05$, $^{**}p<0.01$, $^{***}p<0.001$).}
    \label{fig:goal_infer_six}
\end{figure}



\begin{figure}
    \centering
    \includegraphics[width=\linewidth]{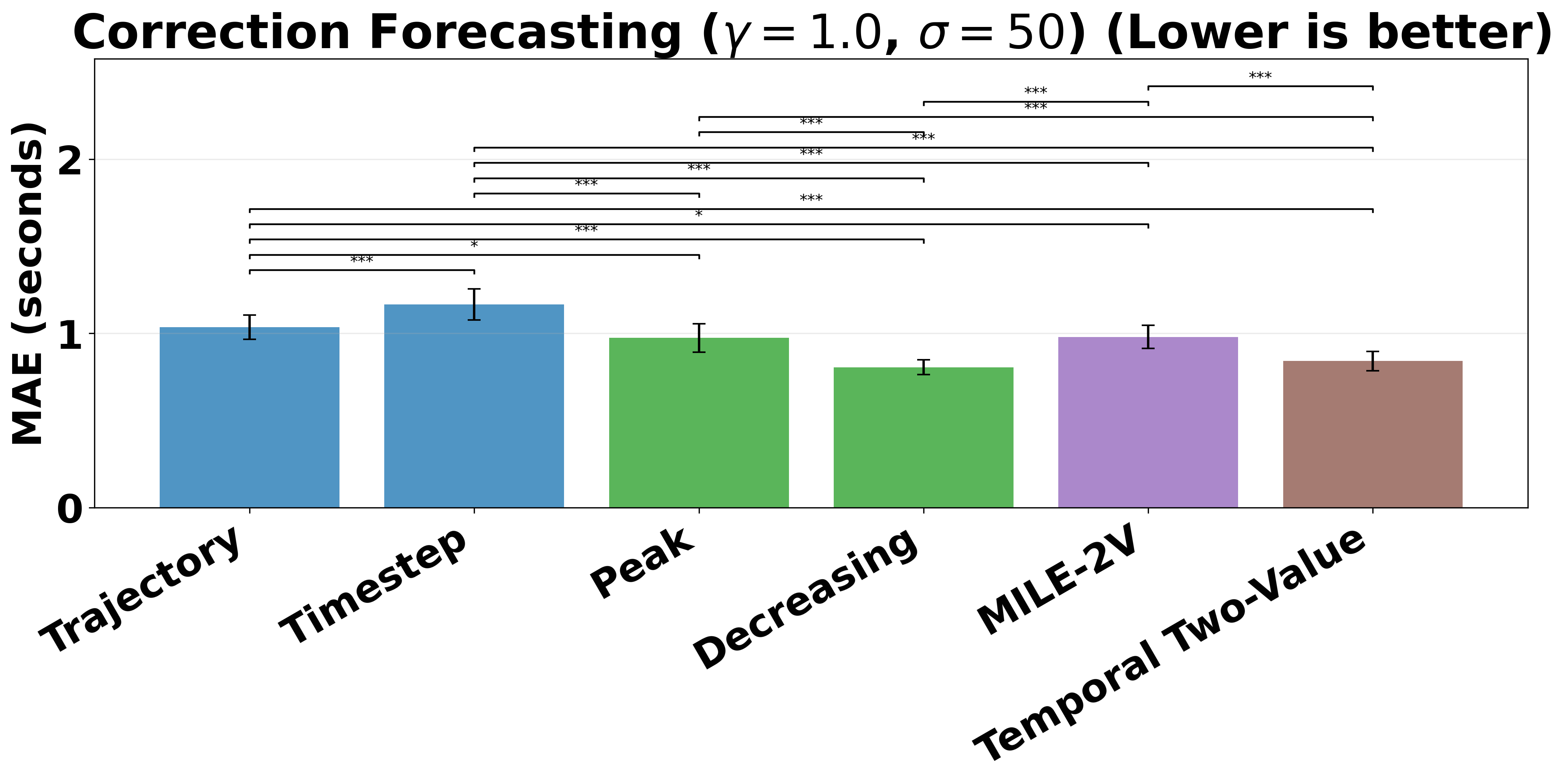}
    \caption{Correction-forecasting performance across the six representative models with $\gamma=1.0$ and $\sigma_h=50$. Bars show mean MAE with 95\% CI (lower is better). Brackets indicate significance ($^*p<0.05$, $^{**}p<0.01$, $^{***}p<0.001$).}
    \label{fig:forecast_six}
\end{figure}

\subsection{Comparison of Preference Formulations}
\label{sec:final_model_comparison}

Based on the preceding model-selection results, we use the selected nonlinear value architecture, neighboring pre-intervention preferences, and forecasting hyperparameters for the final comparison. Temporal Two-Value combines Decreasing for pre-intervention preferences with Timestep for post-intervention preferences, based on their performance across both tasks.

\paragraph{Goal Inference.} Figure~\ref{fig:goal_infer_six} compares the six models on goal inference. Temporal Two-Value achieves the lowest (i.e., best) KL divergence, followed closely by Timestep and Decreasing. Peak-and-Drop and MILE-2V yield higher KL divergence, while Trajectory has the highest mean KL divergence and substantially greater variability. Temporal Two-Value is significantly lower than Trajectory, Peak-and-Drop, and MILE-2V ($p<0.001$), while its differences from Timestep and Decreasing are not significant.

\paragraph{Correction Forecasting.} Figure~\ref{fig:forecast_six} compares the six models on correction forecasting. Decreasing achieves the lowest mean MAE, followed by Temporal Two-Value, Peak-and-Drop, and MILE-2V, while Trajectory and Timestep have higher errors. Temporal Two-Value significantly outperforms Trajectory, Timestep, Peak-and-Drop, and MILE-2V ($p<0.001$); no significant difference is observed between Temporal Two-Value and Decreasing.

\begin{figure}
    \centering
    \includegraphics[width=\linewidth]{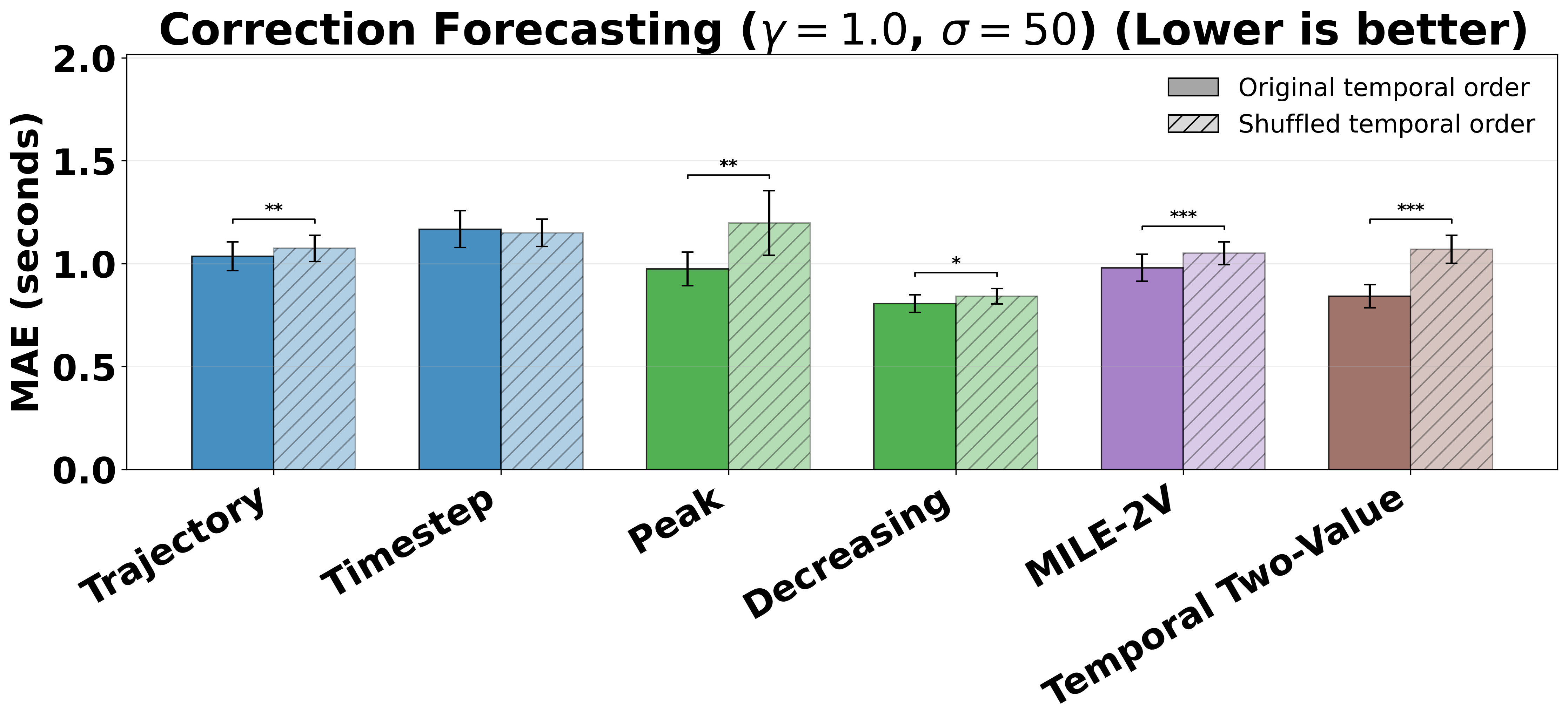}
    \caption{Correction-forecasting performance using original and shuffled value sequences ($\gamma=1.0$, $\sigma_h=50$). Bars show mean MAE with 95\% CI (lower is better). Brackets indicate significance ($^*p<0.05$, $^{**}p<0.01$, $^{***}p<0.001$).}
    \label{fig:forecast_shuffled}
\end{figure}

\subsection{Temporal Randomization Control}
\label{sec:randomization}

Figure~\ref{fig:forecast_shuffled} compares forecasting from the original value sequences with temporally shuffled sequences. Shuffling significantly increases forecasting MAE for Trajectory, Peak-and-Drop, Decreasing, MILE-2V, and Temporal Two-Value, while no significant difference is observed for Timestep. The largest increases are observed for Temporal Two-Value and Peak-and-Drop.

\begin{figure}
    \centering
    \includegraphics[width=\linewidth]{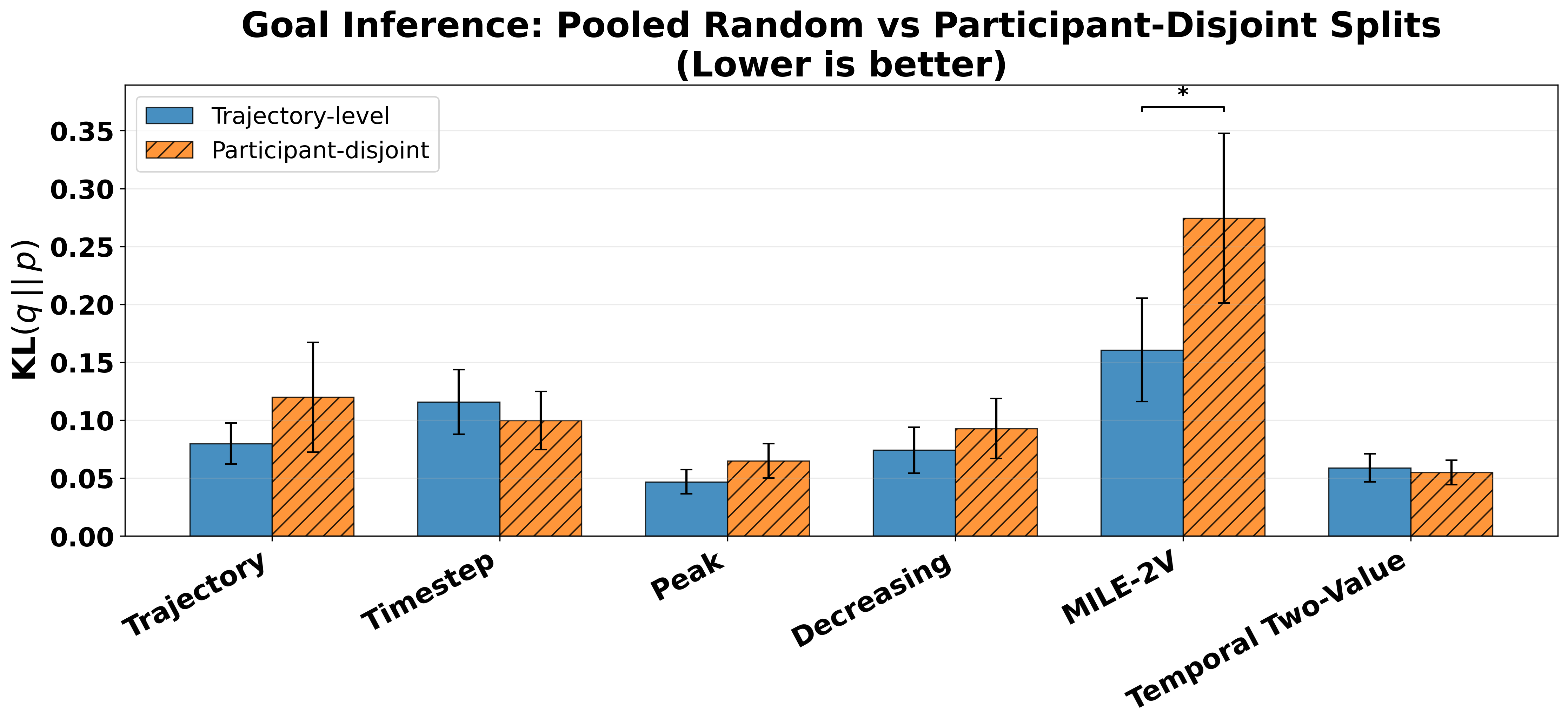}
    \caption{Goal-inference performance under trajectory-level and participant-disjoint splits. Bars show mean KLD with 95\% CI (lower is better). Brackets indicate significant within-model differences ($^*p<0.05$, $^{**}p<0.01$, $^{***}p<0.001$).}
    \label{fig:goal_infer_participant}
\end{figure}

\begin{figure}
    \centering
    \includegraphics[width=\linewidth]{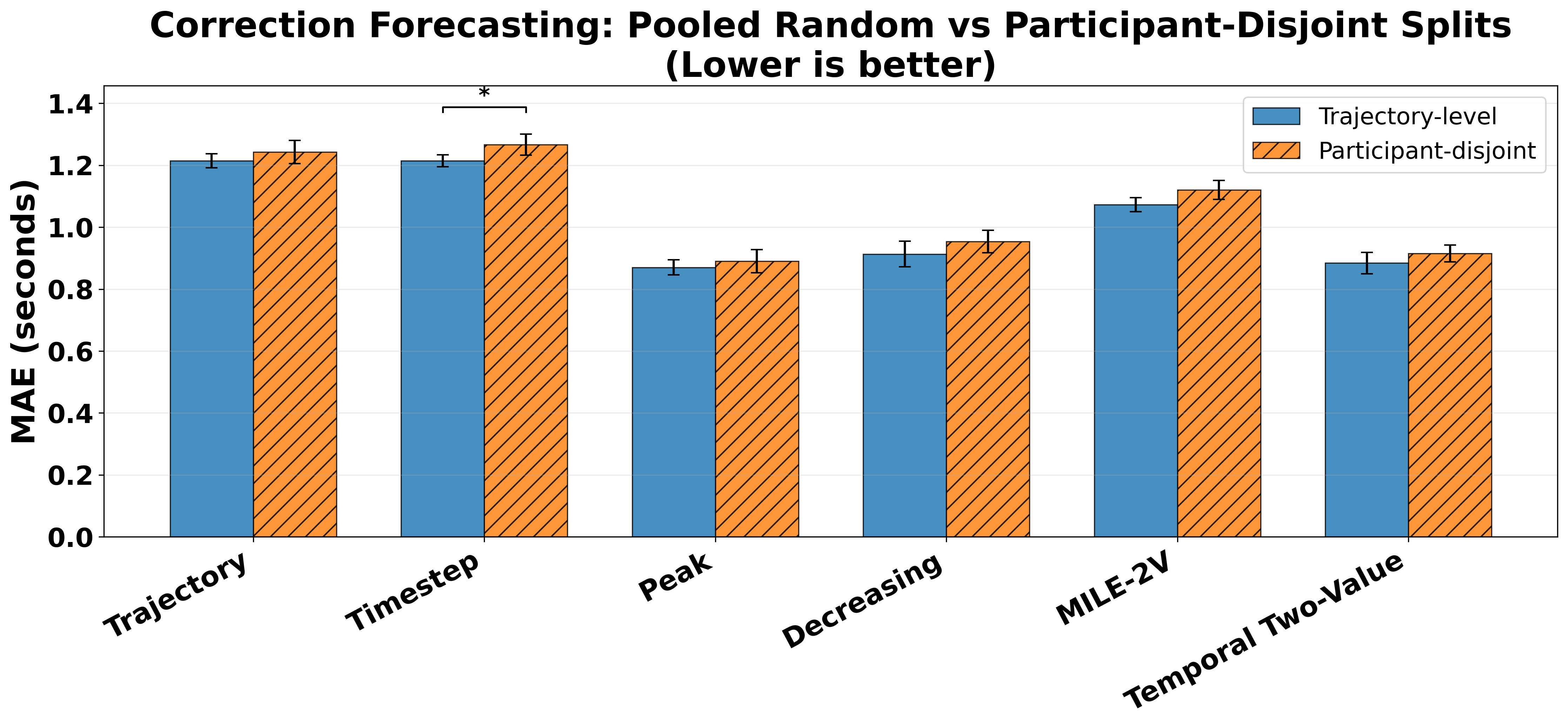}
    \caption{Correction-forecasting performance under trajectory-level and participant-disjoint splits. Bars show mean MAE with 95\% CI. Brackets indicate significant within-model differences ($^*p<0.05$, $^{**}p<0.01$, $^{***}p<0.001$).}
    \label{fig:forecast_participant}
\end{figure}

\subsection{Generalization Across Participants}
\label{sec:participant_generalization}

We finally compare trajectory-level and participant-level generalization using the pooled $0$--$90\%$ dataset described in Sec.~\ref{sec:evaluation_protocol}.

\paragraph{Goal Inference.} Figure~\ref{fig:goal_infer_participant} compares goal inference under trajectory-level and participant-disjoint splits. Performance remains similar across the two splits for most models, with minor but non-significant differences for Trajectory, Timestep, Peak-and-Drop, Decreasing, or Temporal Two-Value. MILE-2V shows significantly higher KL divergence under the participant-disjoint split ($p<0.05$). Temporal Two-Value maintains low KL divergence under both splitting strategies.

\paragraph{Correction Forecasting.} Figure~\ref{fig:forecast_participant} shows a similar pattern for correction forecasting. MAE remains comparable between trajectory-level and participant-disjoint splits for Trajectory, Peak-and-Drop, Decreasing, MILE-2V, and Temporal Two-Value. Timestep shows a significant increase in MAE under the participant-disjoint split ($p<0.05$). Overall, there are limited changes between the two splitting strategies.


\section{Discussion}
\label{sec:discussion}

\textbf{Pre- and post-intervention preferences capture complementary information.} Our results show that when and how a person corrects provide different information about their preferences. Post-intervention Timestep preferences perform particularly well for goal inference, suggesting that corrected behavior provides a strong signal about the human's intended outcome. In contrast, the pre-intervention Decreasing formulation performs particularly well for correction forecasting, where changes in value leading up to an intervention are directly relevant. Thus, behavior before and after a correction captures different aspects of the information communicated through corrective feedback.

\textbf{Local temporal relationships provide stronger preference supervision.} The comparison between pre-intervention formulations suggests that local temporal relationships provide more useful supervision than imposing preferences across all timestep pairs. Neighboring preferences generally outperform all-pairs preferences for both Peak-and-Drop and Decreasing, particularly for goal inference. This suggests that correction timing may provide stronger evidence about local changes in the acceptability of robot behavior, rather than about a complete ordering over the entire preceding trajectory.

\textbf{Forecasting benefits from broader temporal context.} Correction forecasting generally improves with larger $\gamma$ and $\sigma_h$, with the highest tested values providing the strongest performance. A larger $\gamma$ preserves information from earlier values in the history, while a larger $\sigma_h$ gives training samples across a broader range of intervention horizons substantial weight. Together, these results suggest that although local temporal relationships are effective for constructing preference supervision, anticipating when a person will intervene benefits from considering how learned values evolve over a broader temporal context.

\textbf{Combining "when" and "how" preserves information useful across tasks.} Temporal Two-Value achieves the lowest goal-inference KL divergence while remaining among the strongest models for correction forecasting. While individual formulations exhibit task-specific strengths, the combined representation retains information useful for both interpreting the human's intended outcome and anticipating an intervention. This supports treating pre- and post-intervention information as complementary rather than requiring a single source of corrective feedback to capture both.

\textbf{The temporal evolution of learned values carries information about intervention timing.} Shuffling value histories significantly increases forecasting error for five of the six models, including Temporal Two-Value, despite preserving the same underlying values. Thus, forecasting depends not only on value magnitude but also on how values evolve as an intervention approaches. Timestep is the exception, showing no significant degradation under shuffling, consistent with its supervision coming from post-intervention corrective behavior rather than the temporal progression toward an intervention.

\textbf{Preference-based value learning generally transfers across users.} The preference-based formulations show largely consistent performance between trajectory-level and participant-disjoint splits, suggesting that the learned value functions are not strongly dependent on corrections from participants observed during training. In particular, Temporal Two-Value shows no significant degradation under participant-disjoint evaluation for either goal inference or correction forecasting. This contrasts with the MILE-2V baseline, which exhibits significant degradation in goal inference on unseen participants. These results suggest that explicitly translating when and how people correct into preference supervision can support robust cross-participant value learning, with Temporal Two-Value maintaining its performance across both downstream tasks.




\section{Conclusion}

We formulate Learning from Corrections as a preference-based value-learning problem, using relative relationships revealed by corrective interactions without assuming that corrections are optimal demonstrations. Our results show that \emph{when} and \emph{how} humans correct provide complementary supervision: local pre-intervention preferences capture temporal information useful for correction forecasting, while post-intervention preferences support goal inference. Combining both sources performs strongly across tasks and maintains performance on participants unseen during training. Together, these findings show that corrective interactions provide structured preference information for learning human value representations that generalize across users.


\paragraph{Limitations and Future Work.} This work evaluates whether corrections provide sufficient information to learn value representations reflecting human preferences, rather than using these values to directly optimize robot behavior. We therefore evaluate them through goal inference and correction forecasting as proxy tasks. Future work should integrate the learned values into online policy learning and evaluate whether they improve robot adaptation and reduce the need for subsequent corrections during physical interaction.





\bibliographystyle{ACM-Reference-Format}
\bibliography{sample}

\clearpage
\appendix

\section{Feature Details}
\label{sec: features}

\begin{table}[H]
\centering
\small 
\setlength{\tabcolsep}{5pt} 
\caption{Time-series features for modeling robot behavior and human corrections, grounded in prior work linking motion characteristics to human perception and intervention.}
\label{table:features}
\begin{tabular}{    p{0.49\columnwidth}
     >{\centering\arraybackslash}p{0.42\columnwidth}}
\toprule
\textbf{Feature} & \textbf{Equation} \\
\midrule
Expectation Alignment (Vel.) \citep{gopinath2020active,gopinath2022information} & 
$\frac{v_t \cdot v^{\text{opt}}_{t}}{\|v_t\| \|v^{\text{opt}}_{t}\|}$ \\
Expectation Alignment (Pos.) \citep{korkmaz2025mile,tian2023matters} & 
$\| x_t - x^{\text{opt}}_t \|$ \\
Directness Alignment (Vel.) \citep{gopinath2020active,gopinath2022information} & 
$\frac{v_t \cdot v^{g}_{t}}{\|v_t\| \|v^{g}_{t}\|}$ \\
Velocity Consistency \citep{xie2019deep,yao2022imitation} & 
$\frac{v_t \cdot v_{t-1}}{\|v_t\| \|v_{t-1}\|}$ \\
Legibility \citep{dragan2013legibility} & 
$\frac{\int \mathbb{P}(g | \xi_{S \rightarrow x_t}) \gamma_t \, dt}{\int \gamma_t \, dt}$ \\
Task Progress (Dist. to Goal) \citep{gopinath2020active,gopinath2022information} & 
$\| x_t - x_{g} \|$ \\
Optimality \citep{bajcsy2017learning, bobu2018learning, bobu2020quantifying, losey2022physical} & 
$\frac{\exp\left(-\mathcal{L}(\xi_{s \rightarrow x_t})-\mathcal{L}^{\text{opt}}(\xi_{x_t \rightarrow g})\right)}{\exp\left(-\mathcal{L}(\xi_{s \rightarrow x_{t-1}})-\mathcal{L}^{\text{opt}}(\xi_{x_{t-1} \rightarrow g})\right)}$ \\
\bottomrule
\end{tabular}
\end{table}

\subsection{Optimal Trajectories}
\label{sec:opt_pid}

We generate reference trajectories using a PID controller initialized from the robot's current state along the original pre-planned trajectory. These reference trajectories provide the optimal positions \(x_t^{\text{opt}}\) and velocities \(v_t^{\text{opt}}\) used to compute alignment-related features.

At each timestep \(t\), the controller generates a trajectory from the robot's current position \(x_t\) and velocity \(v_t\) toward the goal \(G^*\). We treat this trajectory as an optimal completion because it represents smooth and predictable motion that humans may naturally expect the robot to execute~\citep{maurice2017velocity}. The optimal position and velocity are obtained by interpolating the reference trajectory 0.25 seconds into the future, yielding \(x_t^{\text{opt}}\) and \(v_t^{\text{opt}}\), respectively.

\subsection{Legibility}
\label{sec:legi}

To quantify motion legibility, we adopt the formulation of~\citet{dragan2013legibility}. The legibility score at timestep \(t\) is defined as

\begin{equation}
l_t =
\frac{\int \mathbb{P}(G^* \mid \xi_{S \rightarrow x_t}) \, \gamma_t \, dt}
{\int \gamma_t \, dt},
\label{eq:appendix_legibility}
\end{equation}

where \(\gamma_t = T - t\) weights earlier portions of the trajectory more heavily and \(T\) is the total trajectory duration. The goal posterior is computed as

\begin{equation}
\mathbb{P}(G^* \mid \xi_{S \rightarrow x_t})
\propto
\frac{
\exp\!\left(
-\mathcal{C}(\xi_{S \rightarrow x_t})
-\mathcal{C}(\xi^\text{opt}_{x_t \rightarrow G^*})
\right)
}{
\exp\!\left(
-\mathcal{C}(\xi^\text{opt}_{S \rightarrow G^*})
\right)
}
\mathbb{P}(G^*),
\label{eq:goal_inference}
\end{equation}

which measures how efficiently the observed trajectory prefix \(\xi_{S \rightarrow x_t}\), followed by an optimal completion \(\xi^\text{opt}_{x_t \rightarrow G^*}\), reaches the goal relative to the fully optimal trajectory \(\xi^\text{opt}_{S \rightarrow G^*}\). In our implementation, \(\mathcal{C}(\cdot)\) is defined as the total trajectory length. Higher legibility indicates that the robot's intention to reach \(G^*\) can be inferred earlier from its motion.

\section{Latent Peak Inference Algorithm}
\label{sec: em alg}

For the peak-and-drop model, latent peaks are optimized using the Expectation Maximization (EM)-style procedure in Algorithm~\ref{alg:peak_em}. During peak inference, candidate peaks are restricted to the interval ($[0.2t_c,,0.95t_c]$), and value sequences are smoothed using a moving-average window of size 7. Candidate peaks are required to satisfy a minimum post-peak drop of ($\delta_{\mathrm{drop}}=1.5$), and peak selection is regularized by a temporal penalty ($\lambda_k=0.003$) to discourage peaks occurring excessively early before intervention. During each EM iteration, the value function is trained for 100 epochs using the current peak assignments, with validation evaluated every 5 epochs and the best validation checkpoint retained. We perform up to 20 EM iterations and terminate early when peak assignments stabilize, using a peak-shift tolerance of 0.1 and patience of 2 EM iterations.

\vspace{-0.5em}
\begin{algorithm}[t]
\small
\caption{EM-Style Peak-and-Drop Value Learning}
\label{alg:peak_em}
\begin{algorithmic}

\STATE \textbf{Input:} Goal-conditioned trajectories, train/validation splits, candidate goals $\mathcal G$
\STATE Initialize value function $V$
\STATE Initialize peak assignments $\{t_p\}$ within the candidate peak window

\FOR{$m=1,\ldots,M$}

    \STATE \textbf{M-step: update value function with fixed peaks}

    \FOR{$e=1,\ldots,E$}

        \FOR{each candidate goal $g$ and trajectory}
            \STATE Given current peak $t_p$, construct adjacent peak-and-drop pairs:
            \STATE \hspace{1em} before peak: $(x_{t+1}^g \succ x_t^g)$ for $0 \leq t < t_p$
            \STATE \hspace{1em} after peak: $(x_t^g \succ x_{t+1}^g)$ for $t_p \leq t < t_c$
        \ENDFOR

        \STATE Let $\mathcal D_g=\{(x_a,x_b,y)\}$ denote the preference pairs for goal $g$
        \STATE Compute average preference log-likelihood:
        \STATE \hspace{1em}
        $s(g)=\frac{1}{|\mathcal D_g|}
        \sum_{(x_a,x_b,y)\in\mathcal D_g}
        \log\Phi\!\left(
        \frac{y(V(x_a^g)-V(x_b^g))}{\sigma_p}
        \right)$

        \STATE Compute goal posterior:
        \STATE \hspace{1em}
        $p(g\mid\mathcal D)=
        \frac{\exp(s(g))}
        {\sum_{g'\in\mathcal G}\exp(s(g'))}$

        \STATE Update $V$ by minimizing
        $D_{\mathrm{KL}}(q(g)\,\|\,p(g\mid\mathcal D))$

        \STATE Evaluate validation loss every $n_{\mathrm{val}}$ epochs
        \STATE Save best validation checkpoint within this EM iteration
    \ENDFOR

    \STATE Restore best validation checkpoint for this EM iteration

    \STATE \textbf{E-step: infer latent peaks with updated value function}

    \FOR{each candidate goal $g$ and trajectory}
        \STATE Compute value sequence $V(x_t^g)$ for $t<t_c$
        \STATE Smooth value sequence to obtain $\tilde V(x_t^g)$
        \STATE Define candidate window
        $t \in [\lfloor r_{\min}t_c\rfloor,\lceil r_{\max}t_c\rceil]$
        \STATE Compute post-peak drop
        $\Delta(t)=\tilde V(x_t^g)-\min_{s\in[t,t_c)}\tilde V(x_s^g)$
        \STATE Keep candidates satisfying $\Delta(t)\geq\delta_{\mathrm{drop}}$
        \STATE Select
        $t_p^*=\arg\max_t[\tilde V(x_t^g)-\lambda_k(t_c-t)]$
        among valid candidates
    \ENDFOR

    \STATE Update peak assignments
    \STATE Stop if mean peak shift is below threshold for multiple EM iterations

\ENDFOR

\STATE \textbf{Return:} learned value function $V$ and inferred peak assignments

\end{algorithmic}
\end{algorithm}

\section{Statistical Significance Plots}
\label{app:significance_plots}

\begin{figure*}[t]
    \centering
    \includegraphics[
        width=0.72\textwidth,
        keepaspectratio
    ]{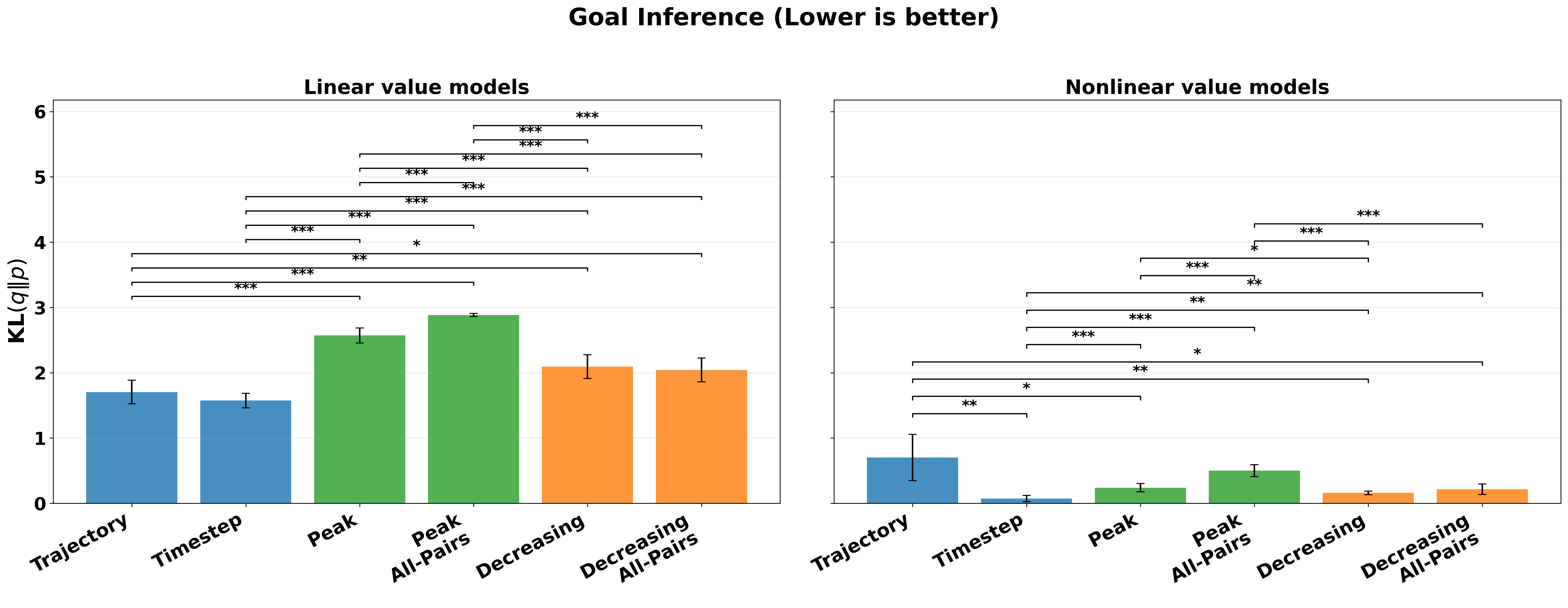}
    \caption{Goal-inference performance across value architectures and preference constructions. Bars show mean KL divergence with 95\% confidence intervals (lower is better). Brackets indicate significant differences ($^*p<0.05$, $^{**}p<0.01$, $^{***}p<0.001$).}
    \label{fig:app_goal_infer}
\end{figure*}

\begin{figure*}[t]
    \centering
    \includegraphics[
        width=0.72\textwidth,
        keepaspectratio
    ]{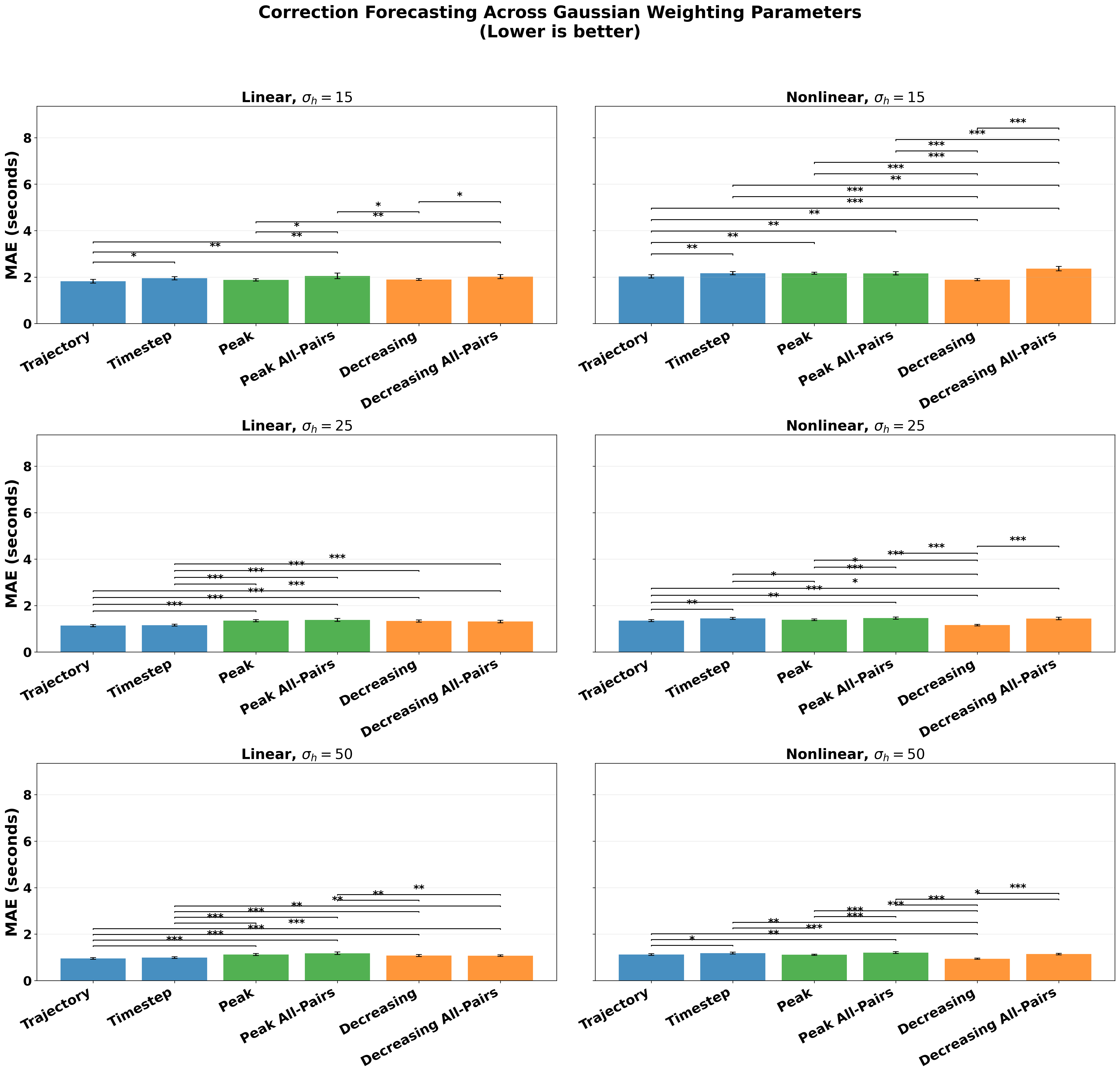}
    \caption{Correction-forecasting performance across Gaussian weighting parameters $\sigma_h\in\{15,25,50\}$. Bars show mean MAE with 95\% confidence intervals (lower is better). Brackets indicate significant differences ($^*p<0.05$, $^{**}p<0.01$, $^{***}p<0.001$).}
    \label{fig:app_forecast_sigma}
\end{figure*}

\begin{figure*}[t]
    \centering
    \includegraphics[
        width=0.6\textwidth,
        keepaspectratio
    ]{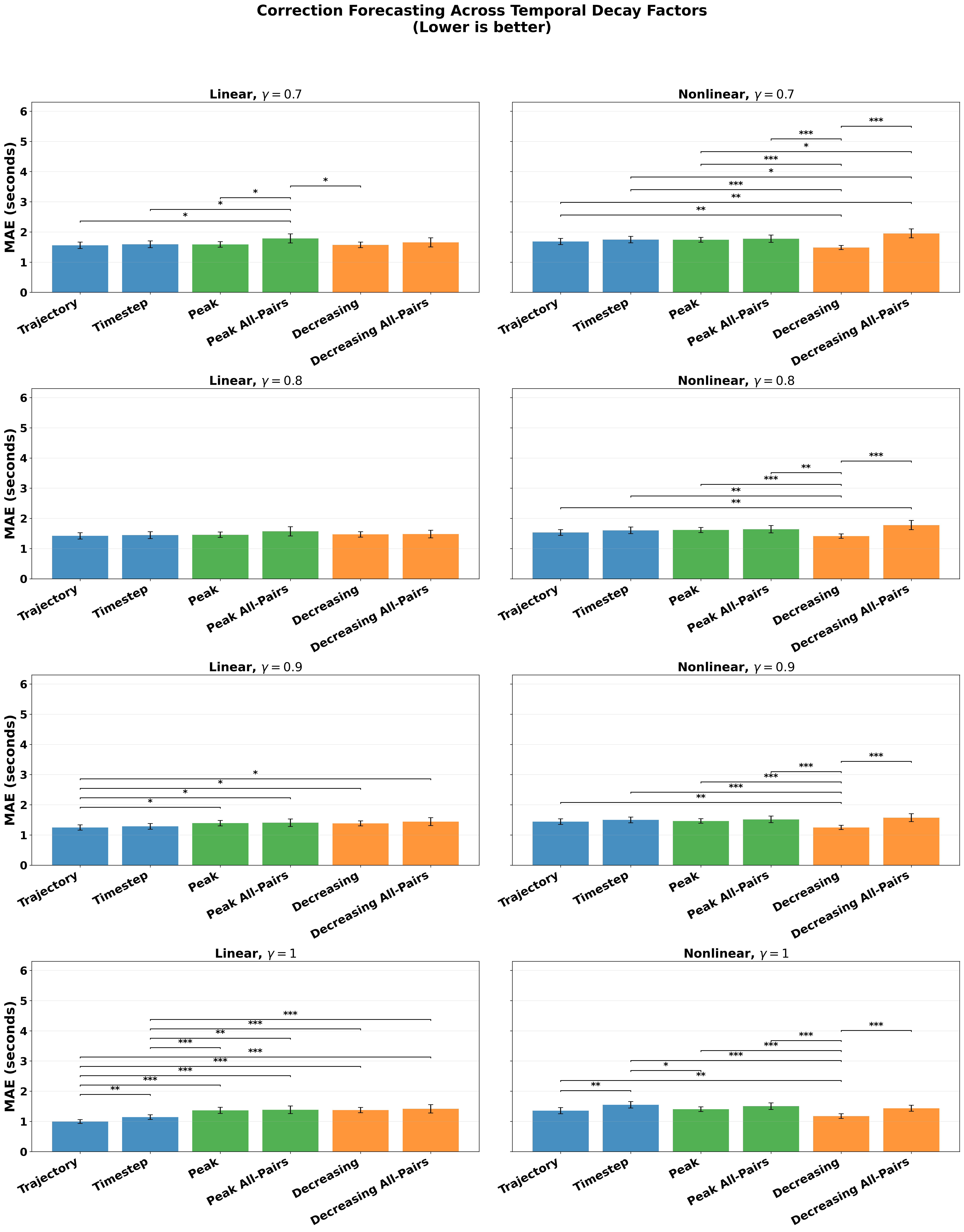}
    \caption{Correction-forecasting performance across temporal decay factors $\gamma\in\{0.7,0.8,0.9,1.0\}$. Bars show mean MAE with 95\% confidence intervals (lower is better). Brackets indicate significant differences ($^*p<0.05$, $^{**}p<0.01$, $^{***}p<0.001$).}
    \label{fig:app_forecast_gamma}
\end{figure*}

Figures~\ref{fig:app_goal_infer}--\ref{fig:app_forecast_gamma} provide the corresponding visualizations of the model- and parameter-selection results reported in Tables~\ref{tab:goal_model_selection}--2, including pairwise statistical significance annotations.

\end{document}